\documentclass[10pt,twocolumn,letterpaper]{article}

\usepackage{cvpr}
\usepackage{times}
\usepackage{epsfig}
\usepackage{graphicx}
\usepackage{amsmath}
\usepackage{amssymb}
\usepackage{float}

\usepackage{float}
\usepackage{enumitem}
\usepackage{booktabs}

\usepackage[pagebackref=true,breaklinks=true,letterpaper=true,colorlinks,bookmarks=false]{hyperref}

\cvprfinalcopy 

\def\cvprPaperID{10}

\ifcvprfinal\fi
\ifcvprfinal\fi
\begin{document}

\title{Recurrent Segmentation for Variable Computational Budgets}

\author{Lane McIntosh\thanks{Work done while an intern at Google Brain.}\\
Stanford University\\
{\tt\small lmcintosh@stanford.edu}
\and
Niru Maheswaranathan\\
Google Brain\\
{\tt\small nirum@google.com}
\and
David Sussillo\\
Google Brain\\
{\tt\small sussillo@google.com}
\and
Jonathon Shlens\\
Google Brain\\
{\tt\small shlens@google.com}
}

\newcommand{\Wih}{\mathbf{W^{ih}}}
\newcommand{\Wix}{\mathbf{W^{ix}}}
\newcommand{\Wfh}{\mathbf{W^{fh}}}
\newcommand{\Wfx}{\mathbf{W^{fx}}}
\newcommand{\Woh}{\mathbf{W^{oh}}}
\newcommand{\Wox}{\mathbf{W^{ox}}}

\newcommand{\Wch}{\mathbf{W^{ch}}}
\newcommand{\Wcx}{\mathbf{W^{cx}}}
\newcommand{\hh}{\mathbf{h}}
\newcommand{\hht}{\mathbf{h}_t}
\newcommand{\xxt}{\mathbf{x}_t}
\newcommand{\cct}{\mathbf{c}_t}

\newcommand{\iit}{\mathbf{i}_t}
\newcommand{\fft}{\mathbf{f}_t}
\newcommand{\oot}{\mathbf{o}_t}
\newcommand{\hhtm}{\mathbf{h}_{t-1}}
\newcommand{\bbc}{\mathbf{b^c}}
\newcommand{\bbh}{\mathbf{b^h}}
\newcommand{\bbi}{\mathbf{b^i}}
\newcommand{\bbf}{\mathbf{b^f}}
\newcommand{\bbo}{\mathbf{b^o}}

\newcommand{\Wxx}{\mathbf{W^{xx}}}
\newcommand{\Wxh}{\mathbf{W^{xh}}}
\newcommand{\Whh}{\mathbf{W^{hh}}}
\newcommand{\Whx}{\mathbf{W^{hx}}}

\newcommand{\Wgy}{\mathbf{W^{gy}}}
\newcommand{\Wyh}{\mathbf{W^{yh}}}
\newcommand{\Wyx}{\mathbf{W^{yx}}}

\newcommand{\Wgyh}{\mathbf{W^{g^yh}}}
\newcommand{\Wgyy}{\mathbf{W^{g^yy}}}
\newcommand{\Wghy}{\mathbf{W^{g^hy}}}
\newcommand{\Wghh}{\mathbf{W^{g^hh}}}
\newcommand{\Wgyx}{\mathbf{W^{g^yx}}}
\newcommand{\Wghx}{\mathbf{W^{g^hx}}}

\newcommand{\xxtm}{\mathbf{x}_{t-1}}
\newcommand{\bbx}{\mathbf{b^x}}
\newcommand{\bby}{\mathbf{b^y}}

\newcommand{\ft}{\text{s}}
\newcommand{\relu}{\text{ReLU}}
\newcommand{\bfg}{b^{fg}}

\maketitle

State-of-the-art systems for semantic image segmentation use feed-forward pipelines with fixed computational costs. Building an image segmentation system that works across a range of computational budgets is challenging and time-intensive as new architectures must be designed and trained for every computational setting. To address this problem we develop a recurrent neural network that successively improves prediction quality with each iteration. Importantly, the RNN may be deployed across a range of computational budgets by merely running the model for a variable number of iterations. We find that this architecture is uniquely suited for efficiently segmenting videos. By exploiting the segmentation of past frames, the RNN can perform video segmentation at similar quality but reduced computational cost compared to state-of-the-art image segmentation methods. When applied to static images in the PASCAL VOC 2012 and Cityscapes segmentation datasets, the RNN traces out a speed-accuracy curve that saturates near the performance of state-of-the-art segmentation methods.
\section{Introduction}

\begin{figure*}
\begin{center}
  \includegraphics[width=0.98\textwidth]{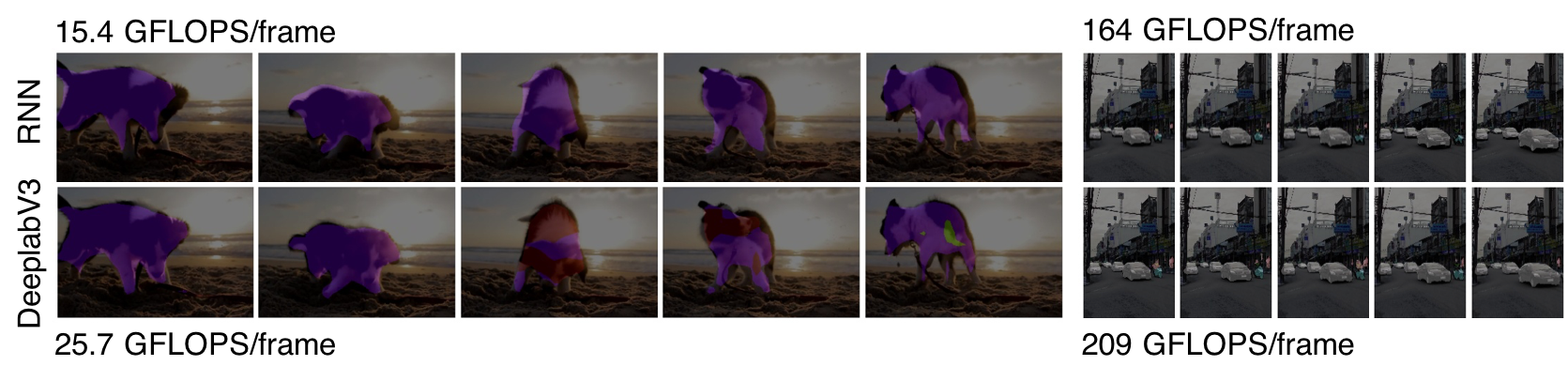}
  \caption{Leveraging a recurrent network to efficiently segment videos. Seeding a recurrent network with the segmentation from a previous frame notably reduces the computational budget required to segment a video.  Segmentations from a recurrent network across $180 \times 270$ and $1080 \times 608$ resolution videos (top row) require about 25\% less computational cost than a state-of-the-art feed-forward architecture \cite{chen2017rethinking} (bottom row), while maintaining higher semantic consistency across frames. Both networks are evaluated at a stride of 16 using a single inference pass.}
  \label{fig:fig1_video}
\end{center}
\end{figure*}

Advances in the design of neural networks have driven the field of computer vision in the last few years \cite{krizhevsky2012imagenet, donahue2014decaf, lecun2015deep}. State-of-the-art artificial vision systems in image recognition \cite{krizhevsky2012imagenet, szegedy2015going, simonyan2014very}, object detection \cite{faster_rcnn, Girshick_2015_ICCV, girshick2014rich, Erhan_2014_CVPR, NIPS2013_5207}, depth estimation \cite{NIPS2014_5539}, semantic segmentation \cite{deeplab, zhao2016pyramid, girshick2014rich, long2015fully, fu2017stacked, ronneberger2015u}, instance segmentation \cite{recurrent_instance, watershed} and many other image processing tasks \cite{toderici2016variable, ledig2016photo} are built on top of novel network architectures.

Designing neural networks is challenging and time-intensive (but see \cite{nasnet}). Additionally, deploying a trained vision system based on a neural network is computationally expensive often requiring dedicated, highly parallelized numerical linear algebra implementations \cite{Tensorflow2015, chetlur2014cudnn, chen2015mxnet} on custom SIMD hardware architectures \cite{jouppi2017datacenter, krizhevsky2012imagenet, chetlur2014cudnn, viebke2015potential}.
In addition to these challenges, applications of vision systems span a wide range of computational budgets--from severe time or energy budgets \cite{selfdriving_cars, howard2017mobilenets, warden_blog,hu2013real} to unconstrained domains  \cite{ronneberger2015u}.

Semantic image segmentation \cite{deeplab, zhao2016pyramid, girshick2014rich, long2015fully, fu2017stacked, ronneberger2015u} is one such task that is commonly employed across problems with vastly different computational demands--from mobile applications \cite{googlepixel2017, barron2015fast, barron2016fast} to medical image diagnosis \cite{ronneberger2015u}.

As larger networks often lead to improved performance \cite{williams2015scaling,nasnet}, a critical aspect of network design is the ability to build systems that operate well across a range of computational budgets--including and especially, extremely small computational budgets \cite{howard2017mobilenets, shufflenet}. 

One may design and train a distinct network architecture for each computational budget \cite{nasnet}, however this approach is time- and labor-intensive and may require the calibration of a cascade-based system \cite{angelova2015pedestrian}. Another approach is to build a fully convolutional network that operates on images of arbitrary size \cite{deeplab}. This approach leads to state-of-the-art models, but at the expense of building models that operate on images downsampled in spatial resolution.

\begin{figure*}[ht]
\begin{center}
  \includegraphics[width=0.98\textwidth]{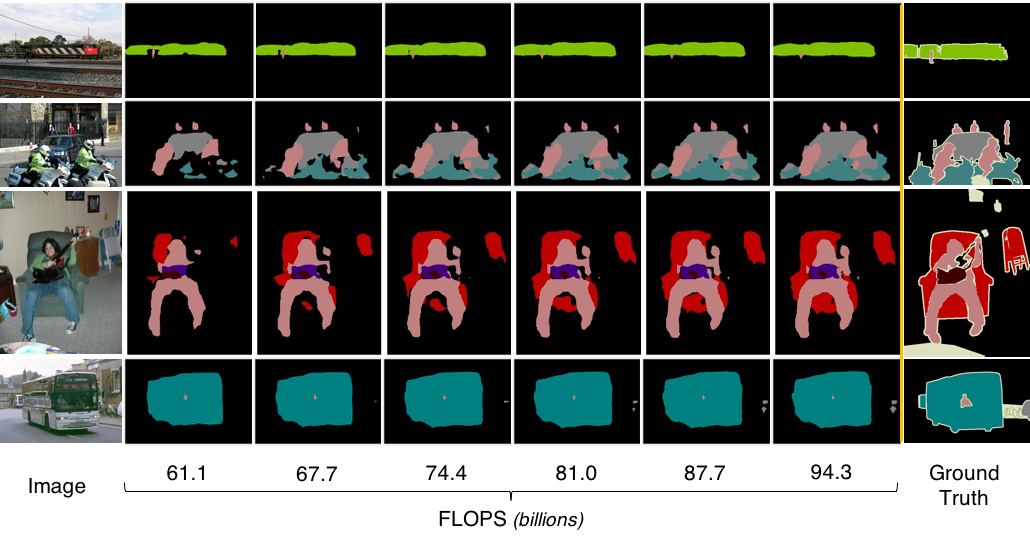}
\caption{Recurrent segmentation iteratively improves segmentation predictions across a range of computational budgets. Images from the PASCAL VOC 2012 \cite{Everingham10} validation dataset at $513 \times 513$ resolution (left column) are segmented with a recurrent network for iterations one through six (columns two through seven), using single-scale evaluation at stride 16 (see Section \ref{sec:methods}). Note that most state-of-the-art segmentation networks use multi-scale inference at significantly higher computational costs. Ground truth segmentations (right column) are shown for comparison. Floating point operations per second (FLOPS) measures the cumulative computational cost of each RNN iteration. The cost at the first iteration includes both a feed-forward ResNet feature extractor and a single pass through the RNN layers, while each subsequent iteration only involves an additional pass through the RNN layers at a marginal cost of less than 7 GFLOPs.}
  \label{fig:fig1_pascal}
\end{center}
\end{figure*}

We instead draw inspiration from the image compression literature, where a single network is often trained and deployed across a large range of computational budgets \cite{toderici2016variable}. A natural way to achieve this goal is through a recurrent network architecture \cite{goodfellow2016deep}, which may be iterated a number of times in proportion to a desired accuracy. Although convolutional networks are far more common in architectures for vision problems, recurrent networks have been applied across tasks such as multiple object recognition \cite{ba2015multiple}, fine-grain image recognition \cite{sermanet2014attention}, instance segmentation \cite{recurrent_instance}, and image generation \cite{gregor2015draw}.

\begin{figure*} 
\begin{center}
  \includegraphics[width=0.9\textwidth]{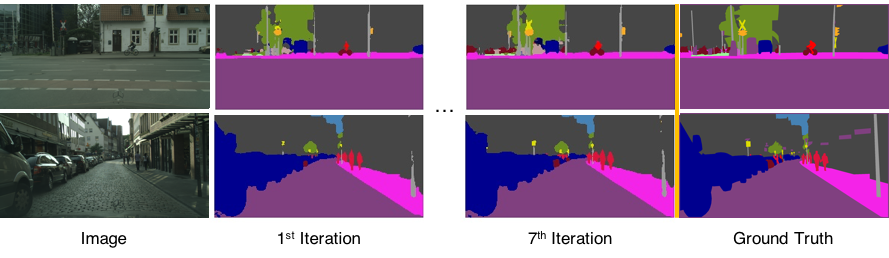}
  \caption{Recurrent segmentation iteratively improves Cityscapes segmentation masks \cite{Cordts2016Cityscapes}. Subsequent iterations add finer spatial details. High resolution images ($1028 \times 2048$) from the Cityscapes semantic segmentation dataset (left column) are segmented by one or seven iterations of a recurrent network (columns two and three). By the seventh iteration, the RNN has added finer spatial segmentations (e.g.  pedestrians, bicyclists, and sign posts) to the previous segmentation of large-scale structures. Ground truth segmentation masks (right column) are provided for comparison.}
  \label{fig:fig1_cityscapes}
\end{center}
\end{figure*}

In this work we propose a recurrent network architecture with an intermediate human-interpretable canvas for semantic image segmentation (Figure \ref{fig:fig1_pascal} and \ref{fig:fig1_cityscapes}). The RNN output at each iteration is added to this canvas, which is then fed in as input to the next iteration. The network thus updates past predictions stored in the canvas to iteratively refine the segmentation. We demonstrate that our RNN formulation may be applied to video with minimal computational cost by seeding the canvas with the segmentation from previous video frames (Figure \ref{fig:fig1_video}). We further demonstrate that our network corrects the errors of a segmentation on subsequent iterations of the recurrent network (Figure \ref{fig:correcting_canvas}). We also visualize the spatial properties of the RNN through time (Figures \ref{fig:fig7_highres} and \ref{fig:spatial_frequency}).

On static images, we demonstrate the performance of the network on the PASCAL VOC 2012 \cite{Everingham10} and Cityscapes \cite{Cordts2016Cityscapes} semantic segmentation datasets. In this domain we find that the RNN architecture allows for more fine-grained control over computational cost than changing the output stride of a convolutional neural network \cite{chen2017rethinking}, but at similar accuracy. One may select the number of iterations in the recurrent network, tracing out a curve of segmentation accuracy versus computational cost (Figure \ref{fig:speed_accuracy}).
\section{Related Work}

Early work on semantic image segmentation learned single layer representations built on top of hand-crafted features (see \cite{deeplab} for review). The resurgence of neural networks enabled learning rich image features for classifying individual pixels, taking into account context and multi-scale information \cite{hariharan2015hypercolumns, dai2015convolutional, eigen2015predicting}. Broadly speaking, many of these systems use convolutional, feed-forward networks to extract deep features at multiple levels (and spatial resolutions) in order to densely predict the label \cite{liu2015semantic, deeplab, noh2015learning}. Two key features of the feed-forward architectures are the use of {\it skip connections} \cite{he2016deep} to bring lower level features into later processing stages \cite{ronneberger2015u} and {\it atrous} or {\it dilated convolutions} \cite{holschneider1990real} to efficiently increase the receptive field size of activation maps without sacrificing spatial resolution \cite{sermanet2013overfeat, deeplab, papandreou2015modeling, giusti2013fast}. 

Instead of relying on a strictly feed-forward architecture, our work focuses on the application of a recurrent network architecture to semantic segmentation. This work draws inspiration from early work applying semantic segmentation to images in small recurrent neural networks \cite{pmlr-v32-pinheiro14}. Pinheiro and Collobert \cite{pmlr-v32-pinheiro14} built convolutional recurrent networks and trained the resulting system on small datasets labeled for pixel-wise segmentation (e.g. Stanford Background, SIFT Flow). In this work, we scale up these networks substantially in depth and overall size. Notably, we use the idea of a human-interpretable canvas that the network continuously (additively) updates during inference as previously used in generative models of images \cite{gregor2015draw, denton2015deep, sukhbaatar2015end}. A human-interpretable canvas allows the network to make a viable prediction at each successive step of the recurrence.

Segmentation systems often use Conditional Random Fields (CRFs) \cite{lafferty2001conditional} as a post-processing step that uses image priors to improve the spatial or temporal consistency of segmentation masks \cite{deeplab, kundu2016feature}. Zheng et al. 2015 \cite{zheng2015conditional} reformulated CRFs as a constrained recurrent neural network (CRF-RNN) that performs approximate variational inference, and found that end-to-end learning of a CRF-RNN on top of a feed-forward convolutional network improved upon previous state-of-the-art results. Here we do not use any CRF post-processing, and provide a generalized recurrent architecture that allows for trading off speed and accuracy.

\section{Methods}
\label{sec:methods}

\begin{figure*}[ht]
\begin{center}
  \includegraphics[width=0.9\textwidth]{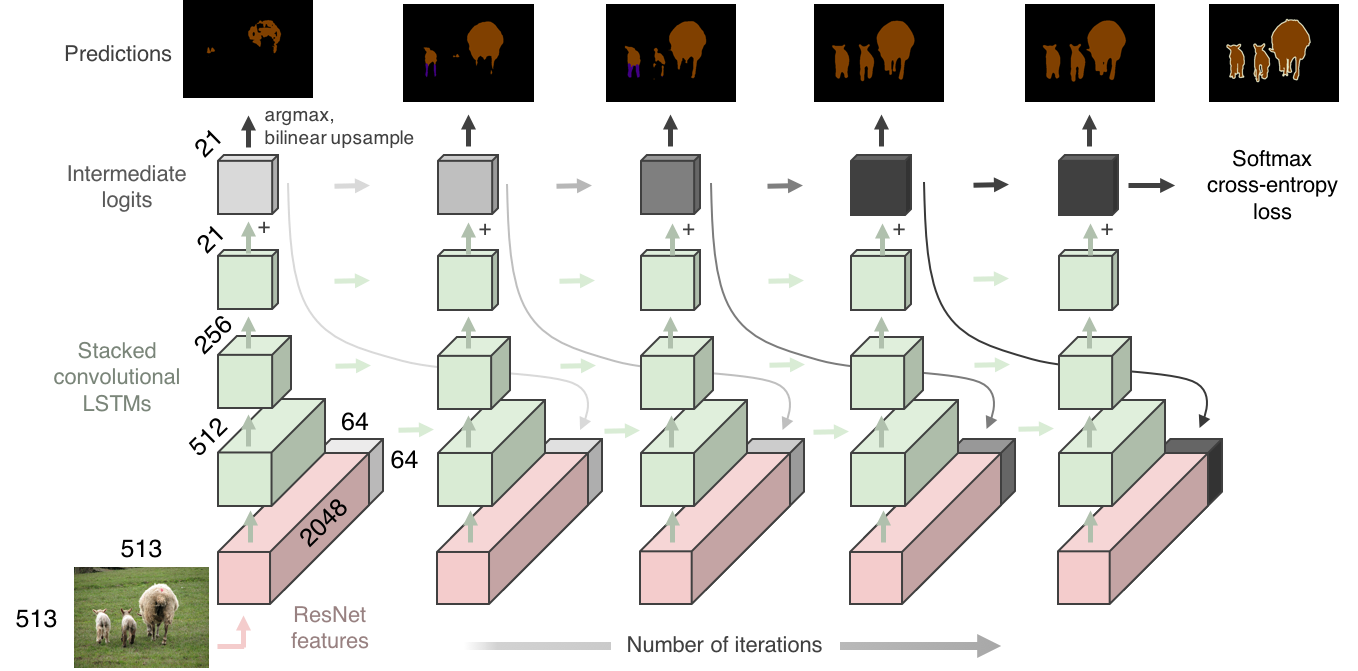}
  \caption{Diagram of recurrent network architecture for image segmentation. The model employs a human-interpretable canvas that is additively updated in subsequent iterations \cite{gregor2015draw}. High-level image features from a pre-trained, fully convolutional network (e.g. \cite{he2016deep, howard2017mobilenets}) are extracted and concatenated with an all-zero ``canvas" which stores the RNN's intermediate logits. On each iteration, a stack of convolutional LSTMs additively updates intermediate logits from the previous iteration by adding their output to the previous iteration's logits. The updated logits are then concatenated with the high-level image features and fed to the RNN at the next iteration.}
  \label{fig:fig4_architecture}
\end{center}
\end{figure*}

The goal of our architecture design is to build an end-to-end fully convolutional recurrent network that may operate on arbitrarily sized input images, where a feed-forward network provides features to a recurrent network. The output of the recurrent network iteratively updates a canvas of semantic segmentation, which is then fed back to the first recurrent layer at the next time step (Figure \ref{fig:fig4_architecture}).  We provide a brief summary of the major features below but save the details for the Appendix.

We compute image features as the output of a conv4\_x (block4) from a ResNet-V1-101 network \cite{he2015deep}. For a $513 \times 513$ image, the image features are of size $65 \times 65$ across 2048 feature channels (red box in Figure \ref{fig:fig4_architecture}). This first portion of the network comprises 42.5M learned parameters and requires 54.4 gigaFLOPS (GFLOPS) to process a single $513 \times 513$ image. Although we experimented with training the convolutional features from scratch, we found that pre-trained features suffice and permit us to accelerate experimentation, hence all experiments presented use ResNet features pre-trained on the COCO semantic segmentation dataset \cite{lin2014microsoft}.

We use LSTM layers \cite{hochreiter1997long} for ease of training \cite{collins2016capacity}. In particular, we use convolutional LSTMs \cite{xingjian2015convolutional} (for definition see Appendix \ref{sec:appendix_methods}). All of the network weights are shared across spatial locations within an activation map.  We used a $1 \times 1$ convolutional kernel size, which combines information only across activation maps. Preliminary experiments found that RNNs with larger than $1\times 1$ kernels--either dense or dilated--underperformed RNNs with $1\times 1$ kernels. In addition to learning the weights and biases during training, we also experimented with learning the initial state of the network. While these experiments produced lower error on the training dataset, they demonstrated poorer generalization; in the results presented here, we initialize the network state as zeros.

Previous work with sequence models \cite{wu2016google} indicate that stacking multiple LSTM layers is a powerful method for capturing dependencies at multiple scales. In preliminary experiments we found that 3 layers provided a reasonable trade off between model complexity and computational cost. Each convolutional LSTM layer maintained the same spatial resolution while decreasing the output depth from the number of image feature channels to the number of semantic classes.

The final output of the network are the logits of an additive canvas \cite{gregor2015draw}. Each iteration of the recurrent network provides additive adjustments to the logits for each class at each pixel location by simply adding the output of the last RNN layer to the canvas. The logits from each iteration are concatenated to the ResNet features to provide input on subsequent iterations of the recurrent network (Figure \ref{fig:fig4_architecture}). The prediction of the network at any iteration is generated by taking the $arg\,max$ of the bilinearly upsampled canvas.

We used a softmax cross-entropy loss across the labels for each pixel. The loss was applied to the final canvas after $N$ iterations, where $N$ is a free parameter discussed below. In preliminary experiments we determined that weighted versions of the loss at earlier iterations of the canvas resulted in similar training performance, but changed the shape of the speed/accuracy curve. In particular, applying the loss to iteration three or earlier improved the prediction quality at earlier iterations, but decreased the quality of the best segmentation and produced a speed-accuracy trade-off curve that did not monotonically increase. All results shown here are for networks trained with a single loss applied at iteration six.

\begin{table*}[ht]
\centering
\scalebox{0.8}{
\setlength{\tabcolsep}{3.2pt}
\begin{tabular}{l|*{20}{c}|c} \toprule
Method & aero & bike & bird & boat & bottle & bus & car & cat & chair & cow & table & dog & horse & mbike & person & plant & sheep & sofa & train & tv & mIOU \\ \midrule \hline
CRF-RNN \cite{zheng2015conditional} & 90.4 & 55.3 & 88.7 & 68.4 & 69.8 & 88.3 & 82.4 & 85.1 & 32.6 & 78.5 & 64.4 & 79.6 & 81.9 & 86.4 & 81.8 & 58.6 & 82.4 & 53.5 & 77.4 & 70.1 & 74.7 \\
DeepLab \cite{deeplab} & 92.6 &  60.4 & 91.6 & 63.4 & 76.3 & 95.0 & 88.4 & 92.6 & 32.7 & 88.5 & 67.6 & 89.6 & 92.1 & 87.0 & 87.4 & 63.3 & 88.3 & 60.0 & 86.8 & 74.5 & 79.7 \\
PSPNet \cite{zhao2016pyramid} & 95.8 & 72.7 & 95.0 & 78.9 & 84.4 & 94.7 & 92.0 & 95.7 & 43.1 & 91.0 & \textbf{80.3} & 91.3 & 96.3 & 92.3 & 90.1 & 71.5 & \textbf{94.4} & 66.9 & 88.8 & 82.0 & 85.4 \\
DeepLabv3 \cite{chen2017rethinking} & 96.4 & 76.6 & 92.7 & 77.8 & \textbf{87.6} & 96.7 & 90.2 & 95.4 & 47.5 & 93.4 & 76.3 & 91.4 & \textbf{97.2} & 91.0 & \textbf{92.1} & 71.3 & 90.9 & \textbf{68.9} & \textbf{90.8} & 79.3 & 85.7 \\
SDN+ \cite{fu2017stacked} & \textbf{96.9} & \textbf{78.6} & \textbf{96.0} & \textbf{79.6} & 84.1 & \textbf{97.1} & \textbf{91.9} & \textbf{96.6} & \textbf{48.5} & \textbf{94.3} & 78.9 & \textbf{93.6} & 95.5 & \textbf{92.1} & 91.1 & \textbf{75.0} & 93.8 & 64.8 & 89.0 & \textbf{84.6} & \textbf{86.6} \\
\hline
RNN$_{\rm{iters}=6}$ & 91.3 & 70.1 & 91.5 & 63.7 & 77.2 & 92.5 & 90.6 & 93.6 & 36.8 & 88.2 & 62.2 & 89.7 & 92.4 & 85.8 & 87.6 & 67.2 & 89.8 & 61.8 & 84.4 & 73.8 & 80.3 \\  \bottomrule
\end{tabular}
}
\vspace{0.2cm}
\caption{\label{tab:pascal_voc_performance} Comparison of the RNN to state-of-the-art image segmentation systems on PASCAL VOC 2012 validation dataset. Performance is measured as intersection-over-union (bold indicates state-of-the-art). RNN segmentations are competitive with best feed-forward architectures. Note the  significant variability across segmentation labels (see Results for discussion). All results are with multi-scale evaluation, and use pre-training on the COCO dataset \cite{lin2014microsoft}.} 
\end{table*}

We trained the resulting architecture on the PASCAL VOC 2012 dataset \cite{Everingham10} which contains 11,530 images of variable size, 6,929 of which have segmentation masks assigning each pixel into one of 20 classes (excluding the background class). We also trained models on the Cityscapes semantic segmentation dataset \cite{Cordts2016Cityscapes}, which consists of 20,000 coarse training images, 3,475 fine validation images and 1,525 test images.

During training, optimization was performed using stochastic gradient descent with momentum, where the learning rate was reduced monotonically with a polynomial schedule (see Appendix for details).

We test the RNN model and competing architectures across \textit{single-scale} and \textit{multi-scale} evaluation schemes. In the single-scale scheme, we perform a single inference pass at a coarse ResNet output stride of 16. This evaluation scheme is generally steered towards mobile platforms with limited computational budgets, and we focus on this in Figure \ref{fig:speed_accuracy}. 
Additionally, we test the model in a less computationally restricted setting by performing multiple inference passes per image: resampling the image at six different spatial scales, performing left-right flips, and averaging the logits, as described in \cite{chen2017rethinking}. The multi-scale setting achieves the best quantitative performance but at a notable computational expense (Table \ref{tab:flops}).

\section{Results}

We present results measuring the performance of the RNN architecture on image segmentation tasks, and highlight experiments indicating how the architecture may be applied to video segmentation in a computationally efficient manner. Finally, we examine the errors of the recurrent network to illustrate how the system operates and suggest opportunities for improvement.

\subsection{Image segmentation with variable computational budgets}
\label{sec:image-segmentation}

We trained the recurrent architecture on the PASCAL VOC 2012 dataset \cite{Everingham10} to segment images into twenty semantic classes. This dataset is a popular test bed for advancements in image segmentation \cite{deeplab, zhao2016pyramid, fu2017stacked}.
The proposed RNN was unrolled for six iterations and a cross entropy loss was applied to the final canvas state (Figure \ref{fig:fig4_architecture}).
The final network is fully convolutional and may operate on arbitrarily-sized images, however we focus our analysis at two operation regimes.

\begin{figure}
\begin{center}
  \includegraphics[width=0.95\columnwidth]{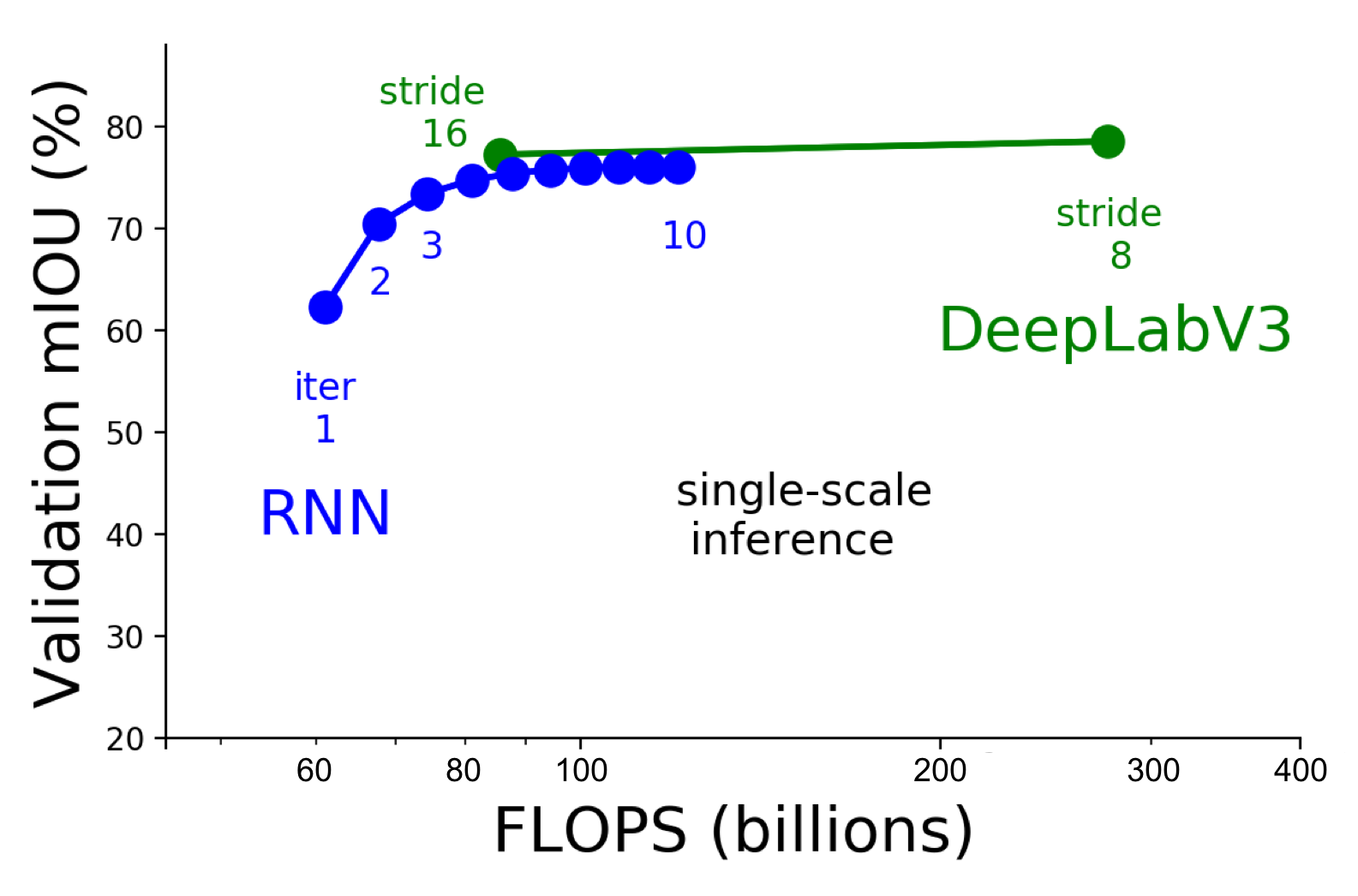}
\caption{Trade-off between computational cost and segmentation accuracy. We plot the computational cost (FLOPS) versus model accuracy (mIOU) on the PASCAL VOC 2012 validation dataset for single-scale evaluation (see Methods). Increasing the number of iterations (``iter'') in the RNN improves model accuracy (blue curves). Additionally, a state-of-the-art fully convolutional network \cite{chen2017rethinking} is evaluated across two output strides (8 and 16). Importantly, recurrent segmentation achieves a finer control over the speed-accuracy trade-off with a similar best-quality segmentation as \cite{chen2017rethinking}. Both networks use the same pre-trained ResNet-101 feature extractor.}
  \label{fig:speed_accuracy}
\end{center}
\end{figure}

First, we analyze the network at a $513 \times 513$ spatial resolution in our single-scale evaluation regime.
A single inference pass of one image at a single iteration results in 61.1 GFLOPs of computation (Figure \ref{fig:fig1_pascal} second column, Figure \ref{fig:speed_accuracy} single-scale, Table \ref{tab:flops}). 
In contrast, state-of-the-art architectures for image segmentation, such as \cite{zhao2016pyramid} and \cite{chen2017rethinking}, require 40\%-100\% more computational cost at the same output stride, indicating that our model is less computationally demanding (Table \ref{tab:flops}).
Interestingly, because of the structure of the RNN, the model may produce intermediate predictions (i.e. by calculating the argmax of the canvas, see Figure \ref{fig:fig4_architecture}). 
Specifically, at inference time, one may continue to iterate the RNN, producing an improved segmentation as the computational budget grows (Figure \ref{fig:fig1_pascal}, third - seventh columns). The computational demand roughly grows linearly with the number of iterations, where each RNN iteration is 10x and 5x cheaper than the comparable post-processing performed in PSPNet \cite{zhao2016pyramid} and DeeplabV3 \cite{chen2017rethinking}, respectively.

\begin{table}
\centering
\scalebox{0.8}{
\setlength{\tabcolsep}{3.2pt}
\begin{tabular}{l|*{4}{c}} \toprule
Method & IOU classes & iIOU classes & IOU categ. & iIOU categ. \\ \midrule \hline
FCN \cite{long2015fully} & 65.3 & 41.7 & 85.7 & 70.1 \\
CRF-RNN \cite{zheng2015conditional} & 62.5 & 34.4 & 82.7 & 66.0 \\
DeepLab \cite{deeplab} & 70.4 & 42.6 & 86.4 & 67.7 \\
PSPNet \cite{zhao2016pyramid} & \textbf{78.4} & \textbf{56.7} & \textbf{90.6} & \textbf{78.6} \\
RNN$_{\rm{iters}=6}$ & 72.7 & 46.0 & 87.6 & 73.9 \\ \bottomrule
\end{tabular}
}
\vspace{0.2cm}
\caption{\label{tab:cityscapes_performance} Comparison of the RNN to state-of-the-art image segmentation systems on the Cityscapes high-resolution validation dataset. Performance is measured as the pixel-wise intersection-over-union (IOU) and instance-level intersection-over-union (iIOU) averaged per category or per class. All models are trained only on the Cityscapes coarse dataset. The RNN consistently achieves second best segmentation despite a reduced computational budget (see Table \ref{tab:flops}).}
\end{table}

We note that fully convolutional networks also have the ability to exchange accuracy for compute, e.g. by decreasing the image resolution or increasing the output stride of the feature extraction step \cite{chen2017rethinking}. When we compare the speed-accuracy trade-offs of changing output stride versus running recurrent segmentation for a variable number of iterations on a $513 \times 513$ image, we find that the RNN achieves a similar quality segmentation when controlling for computational cost, but with more fine grained control over the speed-accuracy trade-off (Figure \ref{fig:speed_accuracy}).

We quantified performance on the PASCAL VOC 2012 test dataset in terms of the mean intersection-over-union (mIOU) between the ground truth labels and network predictions in Table \ref{tab:pascal_voc_performance}. When run for many iterations, the RNN performs comparably to state-of-the-art networks \cite{fu2017stacked,wu2016bridging,deeplab,zhao2016pyramid,chen2017rethinking}, but performs notably worse on specific classes that have high spatial frequency features, such as bicycle and chair. We discuss reasons for this behavior in Section \ref{sec:visualizing} and the Discussion.

We additionally examined how the RNN performs with a very small computational budget. In this case, we computed mIOU versus the number of RNN steps (Figure \ref{fig:speed_accuracy}). The first iteration of the RNN leads to a large performance gain and these gains accrue as one iterates the network. Interestingly, if one runs the network for {\it more steps} than it was originally trained, the RNN predictions continue to improve (Figure \ref{fig:speed_accuracy}, blue curves) as the canvas saturates.

We tested the same network on the Cityscapes \cite{Cordts2016Cityscapes} image segmentation dataset (Table \ref{tab:cityscapes_performance}). 
The performance of the network, trained only on the Cityscapes coarse dataset, was assessed using the mIOU as before -- in addition, we calculated the instance-level intersection-over-union (iIOU), where the contribution of each pixel to the metric is weighted by the ratio of the class' average instance size to the size of the ground truth instance. The RNN provided comparative results with respect to state-of-the-art methods in spite of the reduced computational budget.

\begin{table}
\centering
\scalebox{0.8}{
\setlength{\tabcolsep}{3.2pt}
\begin{tabular}{ll|cc} 
\multicolumn{2}{c}{{\small Method}} & {\small single-scale (GFLOPS)} & {\small multi-scale (GFLOPS)} \\ \midrule \hline
Deeplab V3 \cite{chen2017rethinking} & stride 16 & 85.6 & 1,540 \\
 & stride 8 & 276 & 4,870 \\ \hline
PSPNet \cite{zhao2016pyramid} & stride 16 & 122 & 1,740 \\
 & stride 8 & 260 & 4,090 \\ \hline
RNN & {\it 1} iters & 61.1 & 1,040 \\
 & {\it 2} iters & 67.7 & 1,160  \\
 & {\it 3} iters & 74.4 & 1,269 \\
 & {\it 4} iters & 81.0 & 1,380 \\
 & {\it 5} iters & 87.7 & 1,500 \\
 & {\it 6} iters & 94.3 & 1,610 \\ \bottomrule
\end{tabular}
}
\vspace{0.2cm}
\caption{\label{tab:flops} Computational costs for RNN and state-of-the-art segmentation methods for single-pass and multi-pass evaluation for 513x513 images. Fully convolutional models \cite{zhao2016pyramid, chen2017rethinking} may be evaluated across multiple output strides to reduce computational cost (but with reduced predictive precision). The RNN may be evaluated with an increasing number of iterations to improve predictive performance. Note that the RNN requires a consistently smaller computational budget.}
\end{table}

\subsection{Observations of error correction with recurrent networks}

\begin{figure}
\begin{center}
  \includegraphics[width=1.0\columnwidth]{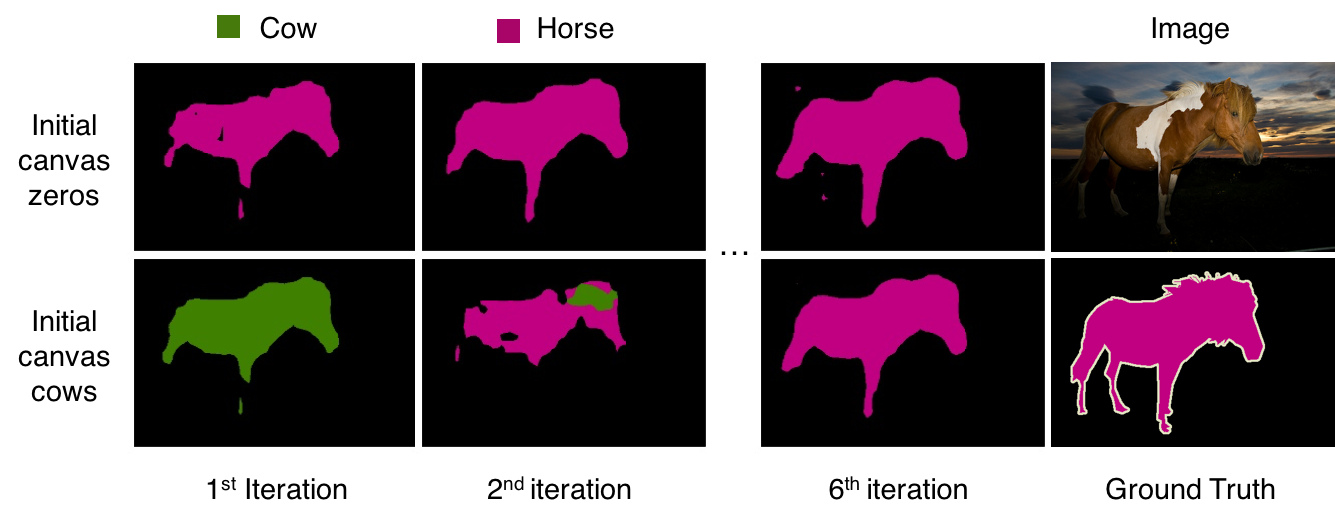}
  \caption{Error correction through RNN dynamics. When the canvas is initialized as zeros, the RNN correctly segments a horse after a single iteration (top row). When the canvas is {\it incorrectly} initialized as a cow, the RNN nonetheless corrects its prediction after two iterations (bottom row).}
  \label{fig:correcting_canvas}
\end{center}
\end{figure}

Given that an RNN performs identical operations on every iteration, it is natural to characterize the computational properties of this generic operation. We addressed this question by artificially perturbing the segmentation mask
stored in the RNN canvas and examining how the RNN responded to these perturbations. 
When we seed the initial canvas to the wrong class--segmenting a horse as a cow, for instance--we find that the RNN is nonetheless able to correct the semantic segmentation in just two iterations (Figure \ref{fig:correcting_canvas}, bottom row). After a few more iterations, the segmentation mask from the perturbed RNN is of similar quality to the RNN initialized with a canvas of all zeros, highlighting the robustness of the method to an incorrect or poor initial segmentation (Figure \ref{fig:correcting_canvas}, top row).

The RNN's ability to iteratively improve partial segmentations (Figure \ref{fig:fig1_pascal}) and correct segmentation errors (Figure \ref{fig:correcting_canvas}) suggests a natural extension to segmenting video. In this setting, consecutive frames are highly correlated and segmentations from previous frames provide a good starting point for subsequent frames.

\subsection{Leveraging recurrent networks for efficient video segmentation}

Temporal correlations between video frames have been exploited in video compression \cite{le1991mpeg} and the propagation of weak labels across video frames \cite{tang2013discriminative, hartmann2012weakly}. In the latter case, optical flow and motion tracking algorithms may learn signals for propagating label information across frames \cite{liu2011sift}.
In the case of semantic video segmentation, we ask whether the structure of the RNN provides a computationally efficient method for label propagation across subsequent frames of a video.

Section \ref{sec:image-segmentation} indicates that subsequent iterations of the RNN refine previous segmentations. A natural question is if the RNN may improve the segmentation provided from a previous frame with only a few iterations, thereby saving significant computational resources. Figure \ref{fig:fig1_video} provides an example pipeline for segmenting recorded video from public domain videos and a mobile device recording, respectively. The first video frame is segmented with the RNN as described above. In subsequent video frames, we seed the RNN canvas with the segmentation from a previous video frame and merely run the RNN for two iterations. The resulting segmentation looks comparable to having run the RNN for many iterations, although the computational demand is significantly reduced. In Figure \ref{fig:fig1_video} we demonstrate that this video segmentation technique is qualitatively comparable to \cite{deeplab} across an array of public domain videos in which ground truth is unknown, at just over half the cost (15.4 GFLOPS/frame for RNN$_{iter=2}$ vs. 25.7 GFLOPS/frame for DeepLab V3 \cite{chen2017rethinking}, on a $180 \times 270$ video).

\subsection{Diagnosing failures of the recurrent network}
\label{sec:visualizing}

\begin{figure}
\begin{center}
  \includegraphics[width=1.0\columnwidth]{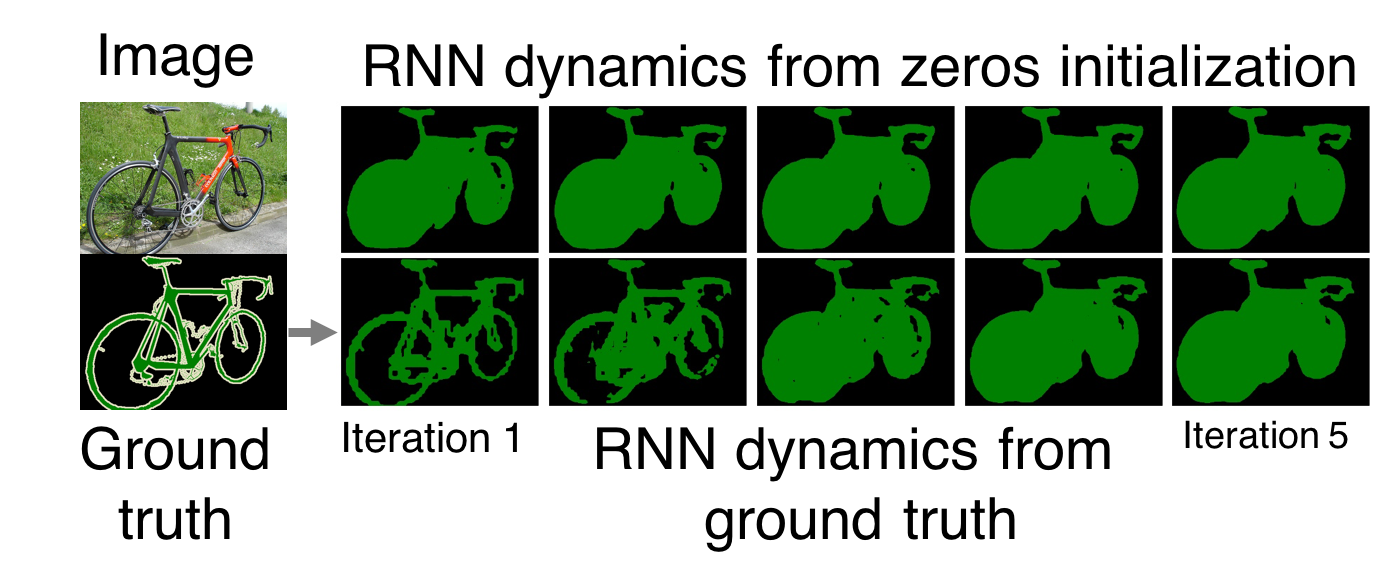}
  \caption{Visualizing RNN dynamics after initializing the canvas with zeros (top row) the ground-truth label (bottom row). Even when the canvas is initialized to the higher spatial frequency ground truth labels, the network fills in pixels in the object's interior. Since the PASCAL VOC dataset segments individual bike spokes and wheels, this represents a failure mode of the RNN.}
  \label{fig:fig7_highres}
\end{center}
\end{figure}

We found that despite achieving near state-of-the-art in several categories of the PASCAL VOC 2012 dataset (Table \ref{tab:pascal_voc_performance}), the performance on several classes--in particular, classes with high spatial frequency features--was poor. To investigate the capacity of the recurrent neural network to learn fine-grained features, we initialized the RNN canvas to the ground truth label of a bicycle with the frame, rims, and wheel spokes all individually segmented (Figure \ref{fig:fig7_highres}, bottom row). When the RNN is allowed to successively iterate on the already correct canvas, the RNN nonetheless progressively fills in the empty areas of the segmentation mask with each iteration, covering a convex hull of the object. The final segmentation mask closely resembles the low spatial frequency mask discovered when the RNN evolves from a blank canvas (Figure \ref{fig:fig7_highres}, top row). We found that even for classes with higher spatial frequencies, the RNN predictions had a spatial frequency distribution that matched the average spatial power spectral density across all classes (Supplemental Figure  \ref{fig:spatial_frequency}).

\section{Discussion}
We propose a recurrent neural network architecture for semantic segmentation that enables variable quality segmentation across computational budgets. This yields a natural framework for segmenting videos, resulting in a high quality segmentation of video frames at a fraction of the computational cost previously feasible.

When applied to images, the RNN model can finely trade-off between accuracy and computational demand. We note however that RNN dynamics tend to fill in the interior of objects with a single semantic label. To address this, it would be possible to target the fine-grained features in these difficult images by bootstrapping \cite{chen2017rethinking} or by explicitly modifying the loss function to penalize object boundaries \cite{watershed}.

Due to memory constraints during training, the RNN processes a relatively coarse spatial resolution due to the output stride applied to the image features. While it is remarkable that the RNN achieves good performance with such coarse-grained features, we anticipate gains in GPU hardware memory
will permit increased spatial resolution
in convolutional RNN architectures and thus allow for more capacity in the network. In the interim, switching from an LSTM to other RNN
architectures with fewer parameters may alleviate some of these memory limitations
during training \cite{goodfellow2016deep}. Preliminary experiments also indicate that pre-training on larger internal datasets \cite{hinton2015distilling} disproportionately improves the performance of the RNN compared to baseline models, resulting in RNNs that achieve state-of-the-art performance on some categories in Pascal VOC. These results suggest that the RNN architecture has a large capacity and may scale to larger problems.

Another direction may be to relax the strict requirement that every iteration of the RNN must perform an identical computation. 
One method for approaching this is to build a hyper-network \cite{ha2016hypernetworks} whereby a second network predicts the network weights for each iteration.
Another approach is to instead build an RNN architecture that adaptively updates the number of iterations \cite{graves2016adaptive}, perhaps on a per-pixel level, in order to achieve a given accuracy level.

The intersection of computational constraints, predictive power, and memory limitations necessitate a diversity of architectural approaches for image and video segmentation. Here we provide one such method, a recurrent architecture that trades off speed and accuracy for semantic segmentation. Developing novel architectures for computer vision problems remains fertile ground for future research.

\section*{Acknowledgements}
We thank Kevin Murphy and George Papandreou for comments on the manuscript, Liang-Chieh Chen for helpful comments about running Deeplab V3, and George Toderici and Ben Poole for useful discussions.

{\small
\bibliographystyle{ieee}
\bibliography{paper}
}

\clearpage

\newpage
\appendix
\section*{Appendix}

\subsection*{Hyperparameters}
\begin{table}[h]
\centering
\begin{tabular}{@{}rllllll@{}} \toprule
Operation      & Kernel size & Stride & Feature maps & Padding &  \\ \midrule
{\bf Network} -- $513 \times 513 \times 3$ input                              \\
ResNet-v1-101  &             &        &              &         &              \\
Convolutional LSTM    & $1$         & $1$    & $512$         &  SAME   &         \\
Convolutional LSTM    & $1$         & $1$    & $256$         &  SAME   &         \\
Convolutional LSTM    & $1$         & $1$    & $|\rm{classes}|$         &  SAME   &         \\
Canvas (add:0)    &         &    &  $|\rm{classes}|$      &    &          \\
Bilinear upsampling     &             &        & $|\rm{classes}|$         &         &              \\
 \midrule
Padding mode           & \multicolumn{6}{@{}l@{}}{Zeros}                    \\
Normalization          & \multicolumn{6}{@{}l@{}}{Batch normalization after every ResNet convolution} \\
Optimizer              & \multicolumn{6}{@{}l@{}}{SGD with Momentum (momentum $=0.95$)}  \\
Parameter updates      & \multicolumn{6}{@{}l@{}}{30,000}                     \\
Learning rate schedule      & \multicolumn{6}{@{}l@{}}{$(1e^{-3} - \epsilon) \cdot
\left (1 - \frac{\mbox{step}}{\mbox{total steps}} \right) ^ {0.9} + \epsilon$} \mbox{ } \mbox{ where $\epsilon = 1e^{-6}$}                    \\
Batch size             & \multicolumn{6}{@{}l@{}}{16}                         \\
Weight initialization  & \multicolumn{6}{@{}l@{}}{Glorot normal \cite{glorot2010understanding}}  \\ \bottomrule
\end{tabular}
\vspace{0.2cm}
\caption{\label{tab:architecture} Details of the recurrent network architecture for image segmentation. $|\rm{classes}|$ is 21 for the PASCAL VOC 2012 semantic segmentation dataset, and 19 for the Cityscapes dataset. The final block4 of the ResNet-v1-101 was augmented with dilation rates of (2, 4, 8) in the three units of block4, following \cite{chen2017rethinking}.}
\end{table}

\clearpage

\subsection*{Supplemental methods}
\label{sec:appendix_methods}
\subsubsection*{Convolutional LSTM}
For the recurrent network architecture, we use stacked convolutional LSTM layers \cite{xingjian2015convolutional} defined as
\begin{align}
  \iit &= \sigma\left(\Wih * \hhtm + \Wix * \xxt + \bbi \right) \\
  \fft &= \sigma\left(\Wfh * \hhtm + \Wfx * \xxt + \bbf + \bfg \right) \\
  \mathbf{c}_t^{in} &= \tanh \left( \Wch * \hhtm + \Wcx * \xxt + \bbc \right) \\
  \cct &= \fft \cdot \mathbf{c}_{t-1} + \iit \cdot \mathbf{c}_t^{in} \\
  \oot &= \sigma\left(\Woh * \hhtm + \Wox * \xxt + \bbo  \right) \\
  \hht &= \oot \cdot \tanh(\cct),
\end{align}
where $\sigma(\cdot)$ is the logistic function, $*$ is the convolution operator, where $i$, $f$, $o$ represent the input, forget and output gates, respectively, and $c$ and $h$ are the state of the LSTM. Forget gate offset bias $b^{fg} = 1.0$.

\subsubsection*{Training details}

We found that the best performance was achieved by having a batch-size of 12 - 16 and a relatively large ResNet output stride of 16. 

We also found that the crop-size had a significant effect on performance, with the highest performance achieved with keeping the crop-size of the input image as large as the native resolution - $513 \times 513$ for PASCAL VOC images and $1025 \times 2049$ for Cityscapes images. In each case the crop size is an integer divisible by 32, plus one, in order to avoid edge effects with the ResNet output stride.

\subsubsection*{Estimating computational cost}

We used the Tensorflow profiler (tf.profiler.Profiler) to estimate FLOPS during evaluation of the models. We also constructed Tensorflow models of the Pyramid Scene Parsing network \cite{zhao2016pyramid} and Deeplab V3 \cite{chen2017rethinking}, following the methods reported as closely as possible. From these models we used Tensorflow profiling as before to estimate the FLOPS for these models. Note however that all performance numbers for both models are taken from the values reported in the original papers.

\begin{figure*} 
\begin{center}
  \includegraphics[width=1.0\columnwidth]{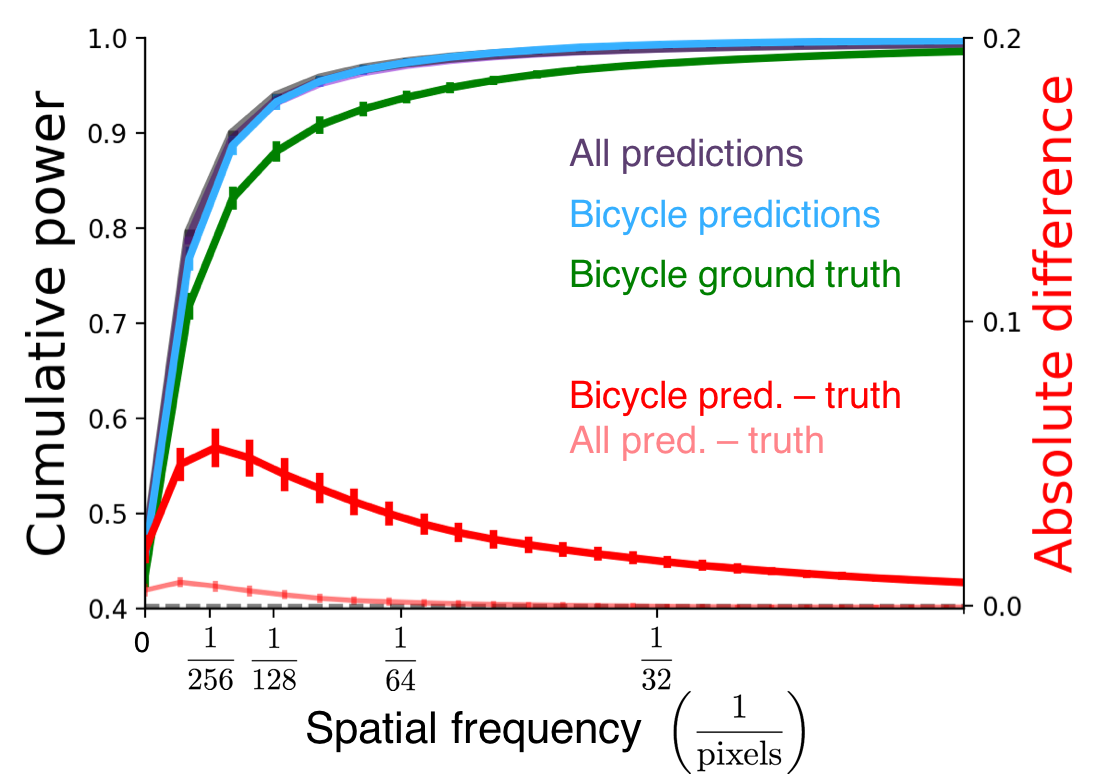}
  \caption{Spatial frequency analysis of segmentation errors. Cumulative distribution of power spectral densities of RNN predictions on all classes (purple), images containing bicycles (blue), and ground truth bicycle segmentations (green). The difference between spectral density distributions for the RNN and ground truth labels are shown in red for the bicycle class (dark red) and all semantic classes (light red). }
  \label{fig:spatial_frequency}
\end{center}
\end{figure*}

\end{document}